\documentclass[twocolumn,10pt]{IEEEtran}
\usepackage{algorithmic,amsbsy,amsmath,amssymb,epsfig,bbm,mathrsfs,multirow,amsthm,subcaption,caption,graphicx,epstopdf,multicol,lipsum,float}
\usepackage{algorithm2e}
\usepackage{enumitem}
\newenvironment{myalign}{\par\nobreak\noindent\align}{\endalign}

\begin{document}

\title{Optimal Stochastic Package Delivery Planning with Deadline: A Cardinality Minimization  in Routing}

\author{\IEEEauthorblockN
{Suttinee Sawadsitang\IEEEauthorrefmark{1},
  Siwei Jiang\IEEEauthorrefmark{2}, 
 Dusit Niyato\IEEEauthorrefmark{1}},
 Ping Wang\IEEEauthorrefmark{1}\\
\IEEEauthorblockA{
\IEEEauthorrefmark{1} School of Computer Science and Engineering, Nanyang Technological University\\
\IEEEauthorrefmark{2}Singapore Institute of Manufacturing Technology (SIMTech) A*STAR } \vspace{-5mm}	}

\maketitle\thispagestyle{empty}

\begin{abstract}
Vehicle Routing Problem with Private fleet and common Carrier (VRPPC) has been proposed to help a supplier manage package delivery services from a single depot to multiple customers. Most of the existing VRPPC works consider deterministic
parameters which may not be practical and uncertainty has to be taken into account. In this paper, we propose the Optimal Stochastic Delivery Planning with Deadline (ODPD) to help a supplier plan and optimize the package delivery. The aim of ODPD is to service all customers within a given deadline while considering the randomness in customer demands and traveling time. We formulate the ODPD as a stochastic integer programming, and use the cardinality minimization approach for calculating the deadline violation probability. To accelerate computation, the L-shaped decomposition method is adopted.  We conduct extensive performance evaluation based on real customer locations and traveling time from Google Map.
\end{abstract}



\section{Introduction}


To achieve a higher profit and  better standard of customer satisfaction, suppliers need to meticulously plan their delivery. The suppliers usually use one of the two delivery modes, which are (i) to deliver customer packages by themselves known as full-truckload (FTL) mode and (ii) to outsource customer package delivery to a third-party carrier known as less-than-truckload (LTL) mode. For the FTL mode, suppliers are required to pay a truck rental, driver stipend, fuel, and other expenses. For the LTL mode, the carriers charge suppliers based on actual package size or weight in an on-demand basic. However, the FTL mode is more economical when the utilization of the FTL trucks is high. The Vehicle Routing Problem (VRP) that combines both FTL and LTL is usually referred to as Vehicle Routing Problem with Private fleet and common Carrier (VRPPC). 

In VRPPC, suppliers need to reserve  FTL trucks in advance before the exact customer demand is observed. The under- and over-reservation of FTL trucks may occur due to unknown customer demand. Furthermore, the suppliers  need to limit their delivery time, i.e., setting the delivery deadline, to save cost, e.g., overtime driver pay and other extra charges. The uncertainty in road traffic, weather, road-construction, and other factors may cause FTL trucks to miss their delivery deadline. To address this issue, one approach is to use a large set of historical customer demand and traveling time which can improve the accuracy of the vehicle routing solution. However, utilizing a large set of historical data incurs more computational time and the problem becomes intractable in practice.
 
In this paper, we aim to help a supplier to minimize their payment of delivering packages from a single warehouse, i.e., a depot, to multiple customer addresses by using either FTL, LTL, or the combination. The delivery has a specific deadline, and missing the deadline will incur a penalty cost to the supplier. Therefore, we propose the Optimal Stochastic Deliver Planning with Deadline (ODPD) to address the aforementioned challenges. The uncertainties in customer demand and delivery time are estimated by historical data. We consider the uncertainty in customer demand in the form of a traditional stochastic integer programming. Alternatively, we consider the uncertainty of traveling time in the form of the cardinality minimization problem. Based on this problem representation, the proposed ODPD does not rely on the strong assumptions, which may not be practical in some cases. These strong assumptions are commonly used in existing works. Such assumptions are (i) the traveling time for each path must follow the normal distribution, (ii) the traveling time of each path must be independent to each other, and (iii) the deadline should be longer than the total traveling time. Moreover, to reduce the computational time of obtaining an optimal solution, we decompose the ODPD into small sub-problems by L-shaped decomposition~\cite{ref_L-shaped}. Finally, the performance evaluation of the ODPD is conducted based on the real dataset obtained from real carrier company and traveling time from Google Map. Compared with the baseline methods, the ODPD can achieve lower deadline violation probability. The execution time of solving the decomposed ODPD is significantly lower than that of the original (nondecomposed) problem. 



\section{Related Work}


 In VRP, a traveling time of a path is random because of road traffic and other factors. The VRP with deterministic traveling time may not be desirable in practical applications. Instead, VRP with random traveling time can be modeled as the stochastic shortest path problem with least expected traveling time (LET) as in~\cite{ref_ev}, \cite{ref_maxpro}, \cite{ref_zhiguang}. The authors in~\cite{ref_ev}  assumed that the weight of each path is independent and follows time-invariant distribution.  They proposed the Expected Value (EV) algorithm and the Expected Lower Bound (ELB) algorithm to solve the problem. 
 From their assumptions, the problem can be regarded as a deterministic shortest path by using the expected values as path weights. Taking different approach, the authors in~\cite{ref_maxpro} proposed a new criterion, which is to reduce the probability of the late arrival instead of to minimize the expected traveling time. This criterion relaxes the problem while keeping the problem practical when people actually plan their journey. 
Later on, the authors in~\cite{ref_zhiguang} adopted the shortest path with deadline violation probability problem from~\cite{ref_maxpro} and  formulated the problem as the cardinality minimization problem. They transformed the cardinality minimization problem into L1-norm minimization and solved it as a mix-integer linear programming. By using cardinality minimization approach, the proposed algorithm can avoid the major assumptions in~\cite{ref_ev} i.e., the weight of each path is independent and follows time-invariant distribution.  
However, the L1-norm minimization does not guarantee to return the optimal result because L1-norm minimization in~\cite{ref_zhiguang} is to minimize the summation of the exceeding time from all samples.

To enhance the quality of business strategies,  many studies extend the shortest path problem by combining it with business planning constraints, e.g. VRPPC, VRP with time window, multi-depot VRP, and other aspects~\cite{ref_greensurvey}. The VRPPC has been earlier introduced in~\cite{ref_1983}. Subsequently, the authors in~\cite{ref_SRI} proposed ``Selection, Routing and Improvement Algorithms" (SRI), which use $\lambda$-interchange  procedure. The authors in~\cite{ref_2opt} proposed the ``Iterated Density Estimation Evolutionary Algorithm (IDEA)", which achieves better computational performance and more accurate results compared to the SRI algorithms. Most of the VRPPC studies were dedicated to improving the quality of the heuristic solver and considered only dynamic parameters. In our previous work~\cite{ref_maggie_vtc}, we considered different aspect of VRPPC, which was to focus on modeling and formulating the VRPPC problem. The  Optimal Delivery Planning (ODP) was  proposed to handle stochastic customer demand. However,  the ODP problem becomes intractable easily  when the number of trucks, carriers, customers, and especially demand scenarios increase. 

For further improvement of the supplier planning, in this paper, we propose the  Optimal Delivery Planning with Deadline (ODPD) as an extension of ODP, where the supplier can limit the delivery time.
In practice,  the uncertainty in traveling time is also unavoidable. 
Thus, we reformulate the system model as the cardinality minimization problem to handle this randomness.  Since the solution of L1-norm formulation in~\cite{ref_zhiguang} does not guarantee to be optimal, we directly minimize the cardinality of all samples instead of minimize the total exceeding time of all samples. 
Furthermore, to speed up the computation, the L-shaped decomposition~\cite{ref_L-shaped} is applied into the ODPD.

\section{System Model and Assumptions}

This section presents the system model for the ODPD. The ODPD is formulated as a two stage stochastic programming model in (\ref{e_obj1}) and (\ref{eq_obj2}). The first stage and second stage decisions are made before and after observing the actual customer demand, respectively. 
\begin{center}
\begin{minipage}{0.45\textwidth}
\textbf{First stage:} The ODPD helps a supplier decide how to reserve FTL trucks based on FTL truck information and customer demand probability. 

\textbf{Second stage:} The ODPD assigns all customers with non-zero demand to either an outsource LTL carrier or the reserved FTL truck. Additionally, the ODPD finds the best delivery routing of the FTL trucks. The routing is calculated based on both traveling distance and the probability of serving all assigned customers before the deadline.
\end{minipage}
\end{center}

The supplier has a set of customers, which is represented as $\mathcal{C} = \{C_1, C_2, \dots, C_{n'}\}$, where $n'$ is the total number of customers. Customer $C_i$ may or may not have demand. The demand of the customer $C_i$ is denoted as $D_i$, where $D_i=1$ if the customer $i$ has demand, and $D_i=0$ otherwise. Let the demand scenario be a collection of all customers' demand represented as $\omega = (D_1, D_2, \dots, D_{n'})$. Let $\Omega = \{\omega_1,\omega_2, \dots, \omega_{q'}\}$ be a set of demand scenarios, where $q'$ is the total number of scenarios.  The list of package weight of each customer is represented as $A = ( A_1, A_2, \dots , A_{n'} ) $.

A set of FTL trucks is represented as $\mathcal{T} = \{T_1, T_2, \dots, T_{t'}\}$ where $t'$ denotes the total number of FTL trucks. Each truck has its own capacity, which is denoted as $F = (F_1, F_2, \dots, F_{t'})$. Let $\mathbb{T}$ denote a deadline or an instant of time that all FTL trucks aim to finish the delivery. The supplier may outsource  third-party LTL carriers to deliver packages. Let $\mathcal{R} = \{R_1, R_2, \dots, R_{r'}\}$ represent a set of LTL carriers, where $r'$ is the total number of LTL carriers. Note that the routing of FTL trucks is determined by the ODPD while LTL carriers are responsible for finding the routing by themselves. The traveling distance and traveling time from location $u$ to location $v$ are represented as $K_{u,v}$ and $Q_{u,v}$, respectively. Without loss of generality, let $u$ and $v$ be taken from the set of customer locations and the supplier depot, which is denoted as $\mathcal{U}$. In this paper, both traveling distance and the probability of deadline violation are used to calculate the best routing of the FTL trucks.

There are four different payments involved in the ODPD, i.e., $\bar{C}_t$ is the initial cost of FTL truck $t$, $\widehat{C}_{i,r}$ is the service charge of LTL carrier $r$ to serve customer $i$, $\ddot{C}_{u,v}$ is the routing cost of FTL truck $t$ from location $u$ to location $v$, and $\acute{C}$ is the penalty cost that incurs when the delivery time exceeds the deadline $\mathbb{T}$.

\begin{table}[t]
\centering
\scriptsize
\caption{List of key notations}
\label{my-label}
\begin{tabular}{c p{6.0cm}}
\hline

\textbf{Symbols} & \textbf{Description} \\ \hline

 \textbf{Main Sets}\\
 $\mathcal{T}$ 		& 	Set of private trucks, where $t \in \mathcal{T}$ is truck's index \\
 						
 $\mathcal{C}$ 		& 	Set of customers, where $i,j \in \mathcal{C}$ are customers' indexes \\
 $\mathcal{U}$			&	Set of routing locations $\mathcal{C} \cup \{depot\}$, where\\& $u,v \in \mathcal{U}$ are locations' indexes\\
 							
 $\mathcal{R}$				&	Set of carriers, where $r \in \mathcal{R}$ is carrier's index \\
 
 \hline
 \textbf{Parameter}\\
 $F_t$			& 	Capacity of truck $t$\\
 $A_i$			& 	Weight of the customer $i$'s package\\
 $D_i$				&	Demand indicator of customer $i$\\
 $K_{u,v}$				&	Distance between locations $u$ and $v$ in kilometer  \\
 $Q_{u,v}$ & Traveling time between locations $u$ and $v$ \\
 $\mathbb{T}$ & Deadline\\
  \hline
 \textbf{Pricing}\\
 $\bar{C}_t$ 	&	Initial cost of FTL truck $t$ \\
 $\widehat{C}_{i,r}$	&	 Service charge of the carrier $r$ that serves customer $i$\\
 				
 $\ddot{C}_{u,v}$	&	Cost of FTL truck $t$ routing from location $u$ to $v$ \\ 
 $\acute{C}$ & Penalty cost of the deadline violation probability\\
  \hline
 \textbf{Uncertainty}\\
 $\Omega$ 	& Set of demand scenarios, where $\omega \in \Omega$ is the scenario's index\\
 $Prob(\omega)$ 	&		Probability of the scenario $\omega$ to happen	\\
  \hline

 \multicolumn{2}{l}{ \textbf{Decision Variables} }\\ 
$X_{i,t}$	&	A binary variable for allocating FTL truck in which $X_{i,t}=1$ if customer $i$ is served by FTL truck $t$, and $X_{i,t}=0$ otherwise\\
$W_t$		&	A binary variable for representing the reservation of truck $t$ in which $W_t=1$ if  FTL truck $t$ is reserved, and $W_t=0$ otherwise \\
$Y_{i,r}(\omega)$		& A binary variable for allocating LTL carrier in which $Y_{i,r}=1$ if customer $i$ is served by LTL carrier $r$, and $Y_{i,r}=0$ otherwise\\
$V_{u,v,t}(\omega)$		& A binary variable that indicates routing path of FTL truck in which $V_{u,v,t}(\omega)=1$ if FTL truck $t$ uses the path from location $u$ to location $v$, and $V_{u,v,t}(\omega)=0$ otherwise\\		
$S_{i,t}(\omega)$	&	An integer variable that assists the system to eliminate sub-tour from the routing solution \\	
	\hline
	
\end{tabular}
\label{t_symbol}
\end{table}


\section{Problem Formulation}
\label{sec:problemformulation}

In this section, we present the original ODPD problem formulation. Then, we reformulate the original ODPD as a cardinality minimization problem.  Here, the total traveling time is from when the FTL vehicle leaves the depot to when it serves all customers and comes back to the depot again.

\subsection{Original ODPD Formulation}

We formulate the ODPD problem as a stochastic integer programming model. In the model, there are five decision variables, i.e., $X_{i,t}$, $W_{t}$, $Y_{i,r}(\omega)$, and $V_{u,v,t}(\omega)$ which are binary variables, and $S_{i,t}(\omega)$ which takes an integer value between one and the total number of customers. 
The description of the decision variables is presented in Table~\ref{t_symbol}.

\vspace{1em}
\noindent Minimize: 
\begin{myalign}
\label{e_obj1}
\begin{split}
& \sum_{i \in \mathcal{C}}\sum_{t \in \mathcal{T}}X_{i,t} 
+ \sum_{t\in \mathcal{T}}\bar{C}_tW_t + E[\mathscr{Q}( X_{i,t}(\omega), W_{t}(\omega) )]	,	
\end{split}
\end{myalign}
{\normalsize where}
\begin{myalign}
\begin{split}
 \mathscr{Q}( X_{i,t}(\omega), W_{t}(\omega) ) = 
 \sum_{i\in \mathcal{C}}\sum_{r\in \mathcal{R}}\sum_{\omega \in \Omega}
Prob(\omega)\widehat{C}_{i,r}Y_{i,r}(\omega) + \\
 \sum_{u\in \mathcal{U}}\sum_{v \in \mathcal{U}}\sum_{t\in \mathcal{T}}\sum_{\omega \in \Omega}	Prob(\omega)\ddot{C}_{u,v}V_{u,v,t}(\omega)	+\\
\sum_{t \in \mathcal{T}}\sum_{\omega \in \Omega} \acute{C} Prob(\omega)Prob(\sum_{u \in \mathcal{U}}\sum_{v \in \mathcal{U}}Q_{u,v}V_{u,v,t}(\omega) \geq \mathbb{T} )
\label{eq_obj2}
\end{split} 
\end{myalign}
subject to: (3) - (14).
\vspace{1em}

The objective function given in (\ref{e_obj1}) and (\ref{eq_obj2}) includes five terms. The first term $\sum_{i \in \mathcal{C}}\sum_{t \in \mathcal{T}}X_{i,t}$ is to minimize the allocation of FTL trucks. The second to fifth terms aim to minimize the total payment which includes (i) the initial cost of FTL trucks, (ii) LTL carrier service charge, (iii) the cost of FTL vehicle routing, and (iv) the penalty cost due to the delivery time exceeding the deadline. Note that the last term in (\ref{eq_obj2}) is to minimize the probability of arriving at the destination after the deadline $\mathbb{T}$. The penalty cost is calculated based on this deadline violation probability. The deadline violation probability of truck $t$ under demand scenario $\omega$ is expressed as $Prob(\sum_{u \in \mathcal{U}}\sum_{v \in \mathcal{U}}Q_{u,v}V_{u,v,t}(\omega) \geq \mathbb{T} )$. $Prob(\omega)$ denotes the probability of demand scenario $\omega$. 

The constraints in (\ref{eq_assign}), (\ref{eq_capacity}), and (\ref{eq_ftl}) are controlling the assignment, limiting the capacity of each truck, and forcing the objective function to include the initial cost of each truck, respectively. Note that $\Delta$ denotes a large scalar number. In addition, (\ref{e_startend}) to (\ref{e_subtour}) are the routing constraints.

\begin{myalign}
&\sum_{t\in \mathcal{T}}X_{i,t} + \sum_{r \in \mathcal{R}}Y_{i,r}(\omega) \geq D_i(\omega), & i \in \mathcal{C}, \omega \in \Omega	\label{eq_assign} \\
&\sum_{i\in \mathcal{C}}A_iX_{i,t} \leq F_t, 	\label{eq_capacity}
& t \in \mathcal{T} \\
&\sum_{i\in \mathcal{C}}X_{i,t} \leq \Delta W_t, 
& t \in \mathcal{T}
\label{eq_ftl}
\end{myalign}

\begin{myalign}
\scriptsize
&V_{u,u,t}(\omega) = 0
&	u\in \mathcal{U}, t\in \mathcal{T}, \omega \in \Omega \label{e_startend}\\
&\sum_{u \in \mathcal{U}} V_{u,0,t}(\omega) \leq 1, 
& t\in \mathcal{T}, \omega \in \Omega \label{e_from0}
\\
&\sum_{u \in \mathcal{U}} V_{0,u,t}(\omega)\leq 1, 
& t\in \mathcal{T}, \omega \in \Omega \label{e_to0}
\\
&\sum_{u \in \mathcal{U}} V_{u,i,t}(\omega)= X_{i,t}D_i(\omega) & i \in \mathcal{C}, t\in \mathcal{T}, \omega \in \Omega \label{e_from}
\\
&\sum_{u \in \mathcal{U}} V_{i,u,t}(\omega) = X_{i,t}D_i(\omega) & i \in \mathcal{C} ,t\in \mathcal{T}, \omega \in \Omega
\label{e_to}
\end{myalign}

\begin{myalign}
&S_{i,t}(\omega) - S_{j,t}(\omega) + |\mathcal{C}|V_{i,j,t}(\omega) \leq |\mathcal{C}|-1,\nonumber \\
& \qquad\qquad\qquad\qquad\qquad 	i, j \in \mathcal{C}, t \in \mathcal{T}, \omega \in \Omega \label{e_subtour} 
\end{myalign}
The last three constraints in (\ref{e_bound1})-(\ref{e_bound3}) indicate the types and bounds of the decision variables. 

\begin{myalign}
&X_{i,t},W_{t} \in \{0,1\},\qquad\qquad\qquad \qquad
 i \in \mathcal{C}, t \in \mathcal{T} \label{e_bound1}\\
&Y_{i,r}(\omega),V_{u,v,t}(\omega), \in \{0,1\}, \nonumber \\
&\qquad\qquad\qquad r \in \mathcal{R}, u, v \in \mathcal{U}, t \in \mathcal{T}, \omega \in \Omega \label{e_bound2}\\
&S_{i,t,\omega} \in \{0,1,2,3,\dots,n'\}, i \in \mathcal{C}, t \in \mathcal{T}, \omega \in \Omega \label{e_bound3}
\end{myalign}

\subsection{Reformulate the system fomulation with cardinality}

While the probability of demand scenario is a given parameter, the deadline violation probability is calculated as expressed in the cardinality minimization formulation. In particular, the cardinality is the number of non-zero elements in a vector or a matrix. For example, for $\vec{x} = \begin{bmatrix} 0 & 3 & 4& 0 &8& 1 \end{bmatrix}$, the cardinality is $Card(\vec{x}) = 4$.

To reformulate the optimization as the cardinality minimization problem, a set of traveling time samples is required. Let $S =\{S_1,S_2,\dots, S_{s'}\}$ denote a set of traveling time samples, where $s'$ is the total number of samples. The set of traveling time from location $u$ to location $v$ in the sample $\{S_1,S_2,\dots, S_{s'}\}$ is denoted as $\{W_{u,v}^1,W_{u,v}^2, \dots , W_{u,v}^{s'} \}$. Let $G_t(\omega) = \{g_{1,t}(\omega),g_{2,t}(\omega),g_{3,t}(\omega),\dots , g_{s',t}(\omega)\}$ be a set of the time exceeding the deadline that truck $t$ uses to deliver packages in scenario $\omega$. Note that $V_{u,v,t}(\omega)$ is the decision variable. For traveling time sample~$s$, we let $g_{s,t}(\omega)$ equal zero when the delivery is done before the deadline $\mathbb{T}$, and the formulation is expressed as follows:
\begin{myalign}
& g_{s,t}(\omega) = \max \left(0, \left( \sum_{u\in \mathcal{U}}\sum_{v \in \mathcal{U}}W_{u,v}^sV_{u,v,t}(\omega) \right) -\mathbb{T} \right)	.
\label{e_card_g} 
\end{myalign}

Therefore, the deadline violation probability can be calculated by the cardinality of the time exceeding the deadline divided by the total number of samples as follows:
\begin{myalign}
&Prob	\left( \sum_{u \in \mathcal{U}}\sum_{v \in \mathcal{U}}Q_{u,v}V_{u,v,t}(\omega) \geq \mathbb{T} \right) = \dfrac{ Card(G_t(\omega))}{s'}	.
\label{e_card_scount} 
\end{myalign}
Note that the denominator $s'$ is a constant, and we can simplify the fraction expression by including the denominator in the penalty cost~$\acute{C}$.

We introduce a new decision variable for reformulating the ODPD as the cardinality minimization problem. $Z_{t,s}$ is a variable indicating the deadline status of truck $t$ associated with traveling time sample $s$ in which $Z_{t,s} = 1$ if FTL truck $t$ will not be able to deliver packages to all assigned customers before deadline $\mathbb{T}$, and $Z_{t,s} = 0$ otherwise. We introduce the expression
\begin{myalign}
	&Card(G_{t}(\omega)) = \sum_{s\in S}Z_{t,s}(\omega) \label{eq_card_obj}
\end{myalign}
as a part of the objective function. Correspondingly, the constraint in (\ref{eq_card_constraint}) must be introduced together with (\ref{eq_card_obj}).  The constraint in (\ref{eq_card_constraint}) is to force $Z_{t,s}(\omega) = 1$ in which truck $t$ takes longer time to deliver than the deadline $\mathbb{T}$ when using traveling time sample $s$. 
\begin{myalign}
& \sum_{u \in \mathcal{U}}\sum_{v \in \mathcal{U}}W_{u,v}^sV_{u,v,t}(\omega) - \Delta Z_{t,s}(\omega) \leq \mathbb{T}, \nonumber\\
& \qquad\qquad\qquad\qquad\qquad\qquad t\in\mathcal{T}, s\in S, \omega \in \Omega \label{eq_card_constraint}
\end{myalign}
The constraint in (\ref{e_bound4}) indicates that $Z_{t,s}$ is a binary variable.
\begin{myalign}
& Z_{t,s}(\omega) \in \{0,1\},\qquad\qquad s \in \mathcal{S}, t \in \mathcal{T}, \omega \in \Omega \label{e_bound4}
\end{myalign}


\section{Decomposition}

When the number of demand scenarios is large, the optimization problem presented in Section~\ref{sec:problemformulation} becomes more complicated and takes long time to be solved. To make the optimization problem more computationally tractable, we apply L-shaped decomposition method~\cite{ref_L-shaped}. We first reformulate the optimization problem into a master problem and a sub-problem. Since the ODPD is a two stage stochastic optimization problem, the master problem and the sub-problem are proposed to solve the first stage and the second stage, respectively. Both master and sub-problems are solved iteratively in a loop until the optimal solution is found.

\subsection{Master Problem}

The master problem is expressed as follows:

\noindent Minimize: 
\begin{myalign}
\label{e_masterobj}
\begin{split}
& \sum_{i \in \mathcal{C}}\sum_{t \in \mathcal{T}}X_{i,t} 
+ \sum_{t\in \mathcal{T}}\bar{C}_tW_t + \theta	,	
\end{split}
\end{myalign}
subject to: (3) - (4), (12), (21).

The objective function of the master problem is expressed in (\ref{e_masterobj}), where $\theta$ is an additional positive continuous decision variable. The constraints on the capacity limit and the initial cost of FTL trucks are included in the master problem. The optimal solution values obtained from the master problem are denoted as $\bar{X}_{i,t}$, $\bar{W}_{t}$, and $\bar{\theta}$. In the traditional L-shaped method, the additional constraints consist of feasibility cuts and optimal cuts. In this paper, we present only the optimal cut constraint because the feasibility cuts are not necessary. This is from the fact that the first stage decision variables of the ODPD are binary variables. The optimal cut constraint is given as follows:
\begin{myalign}
	&	\sum_{i \in \mathcal{C}}\sum_{t \in \mathcal{T}}E_{i,t}X_{i,t} + \theta \geq e,
\label{e_cut}
\end{myalign}
which requires a solution from the sub-problem, i.e., $E_{i,t}$ and~$e$.

\subsection{Sub-problem}

The sub-problem is given as follows:

\noindent Minimize: 
\begin{myalign}
\begin{split}
& \sum_{i\in \mathcal{C}}\sum_{r\in \mathcal{R}}
Prob(\omega)\widehat{C}_{i,r}Y_{i,r}  + \sum_{t \in \mathcal{T}}\sum_{s \in \mathcal{S}} Prob(\omega) \acute{C}Z_{t,s}\\
& + 
 \sum_{u\in \mathcal{U}}\sum_{v \in \mathcal{U}}\sum_{t\in \mathcal{T}}	Prob(\omega)\ddot{C}_{u,v}V_{u,v,t}	
\label{e_subobj}
\end{split} 
\end{myalign}

\noindent subject to: (3), (6)-(11), (13)-(14), (18)-(19).

The objective function of the sub-problem is to minimize the expected value of the second stage. In the sub-problem, in constraint (3), the decision variable $X_{i,t}$ becomes the parameter denoted as $\bar{X}_{i,t}$, the value of which is obtained from the master problem. Note that $W_{t}$ does not appear in any constraints in the sub-problem, and thus $\bar{W}_{t}$ is not used to solve the sub-problem.

\subsection{Algorithm}

The L-shaped decomposition algorithm for the ODPD is presented in Algorithm~\ref{al_1}, where $k$ is the iteration index. In the first iteration, the master problem is solved by ignoring the optimal cut constraint, and $\bar{\theta}$ is set to a very small value. For the later iterations, the optimal cut constraint is added, and $\theta$ becomes a decision variable, the optimal solution of which is denoted as $\bar{\theta}$. Next, the second stage solutions are solved for all scenarios as the inner loop.

At the end, the values of  $e$ and $E_{i,t}$ are calculated by adapting from the original L-shaped method because the ODPD is a mix integer programming, which is not solvable by simplex method.
The value of  $e$ depends on the values of $J(\omega)$, $P(\omega)$, and  $M(\omega)$, which are the LTL cost for outsourcing all package delivery, the penalty cost, and the routing cost, respectively. 
To calculating $E_{i,t}$, we use similar idea from  $J(\omega)$ and $M(\omega)$, but the value is specific for each $i \in \mathcal{C}$ and $t \in \mathcal{T}$.

When the difference between $\bar{\theta}$ and $B$ is less than $\epsilon$, i.e., error tolerance, the convergence condition is met.
The loop is terminated when the total cost $H$  increases and the convergence condition has been met before. The optimal solution is achieved at the second latest iteration, which is the iteration before the total cost increases.

\begin{myalign}
&P(\omega) = Prob(\omega)\sum_{t\in\mathcal{T}} \sum_{s \in \mathcal{S}}\acute{C}\bar{Z}_{t,s} 
\label{eq_al_penalty}\\
& M(\omega) =  Prob(\omega)\sum_{i\in\mathcal{C}}\sum_{t \in \mathcal{T}}\sum_{u \in \mathcal{U}}\left( \ddot{C}_{u,i}\bar{V}_{u,i,t} +\ddot{C}_{i,u}\bar{V}_{i,u,t} \right)
\label{eq_al_routing}\\
& J(\omega) = Prob(\omega) \sum_{i\in \mathcal{C}} D_i(\omega) \min\limits_{r \in \mathcal{R}}\widehat{C}_{i,r}
\label{eq_al_j}\\
& I_{i,t}(\omega) = Prob(\omega)\left( D_i(\omega) \min\limits_{r \in \mathcal{R}}\widehat{C}_{i,r} \right)  \nonumber\\
& \hspace{3em} -Prob(\omega)\sum_{u \in \mathcal{U}}\left( \ddot{C}_{u,i}\bar{V}_{u,i,t} +\ddot{C}_{i,u}\bar{V}_{i,u,t} \right)
\label{eq_al_i} 
\end{myalign}

\begin{algorithm}[]
\SetAlgoLined

\KwResult{The optimal solution is achieved at the second latest iteration }

 \While{ ($H^{k} < H^{k-1}$) or ($N \neq 1$)}{
   
 \eIf{$k=0$}
 {
 {\ttfamily //first iteration}\;
 - $\bar{\theta}^k$ is set to a very small value\; 
 - Solving master problem without the additional cutting constraint\;
 }{
 - Generating the additional constraint as in (21) by using $e^{k-1}$ and $E^{k-1}_{i,t}$ \;
 - Solving master problem with the additional cutting constraint\;
 }
 - Obtaining all $\bar{X}^k_{i,t}$, $\bar{W}^k_{t}$ and $\bar{\theta}^k$\;

 \ForEach{ scenario $\omega \in \Omega$}{%
 - Solving the sub-problem by using $\bar{X}_{i,t}^k$ as parameters\;
 - Obtaining $\bar{Y}_{i,r}^k$, $\bar{V}_{u,v,t}^k$, and $\bar{Z}_{t,s}^k$\;
 - Calculating  $P(\omega)$ , 
 $M(\omega)$, and $J(\omega)$ 
 from Equation~(\ref{eq_al_penalty}), (\ref{eq_al_routing}), and (\ref{eq_al_j}), respectively\;
 - Calculating $I_{i,t}(\omega), \forall t \in \mathcal{T}, \forall i \in \mathcal{C}$ from (\ref{eq_al_i})\;
 - Calculating $h(\omega)$, which is the objective cost~(\ref{e_subobj}) by using $\bar{Y}_{i,r}^k$, $\bar{V}_{u,v,t}^k$, and $\bar{Z}_{t,s}^k$ as parameters\;
 }
 
 - Calculating $e^{k} =  \sum_{\omega \in \Omega}\left( J(\omega) - P(\omega) -M(\omega)\right)$\;
  $E^{k}_{i,t} =\sum_{\omega \in \Omega}I_{i,t}(\omega) , \forall t \in \mathcal{T}, \forall i \in \mathcal{C}$\;
$B = \sum_{\omega \in \Omega}\left(J(\omega) -  M(\omega)\right)-\sum_{i \in \mathcal{C}}\sum_{t \in \mathcal{T}} E^{k}_{i,t}\bar{X}_{i,t}^k $\;
$H^k = \sum_{t \in \mathcal{T}}\bar{C}_t\bar{W}_t^k + \sum_{\omega \in \Omega}h(\omega)$ \;

 \If{ $|B - \bar{\theta}^{k}|   > \epsilon$}
 {
  $N = 1$ \tcp*{ convergence condition is met }
 } 
  - Increasing iteration index by one $k = k+1$\;
 
 }
 \caption{Decomposition algorithms}
 \label{al_1}
\end{algorithm}
\begin{figure*}[t]
$\begin{array}{ccc} 
\hspace{-1.5em}
\includegraphics[width=0.34\textwidth]{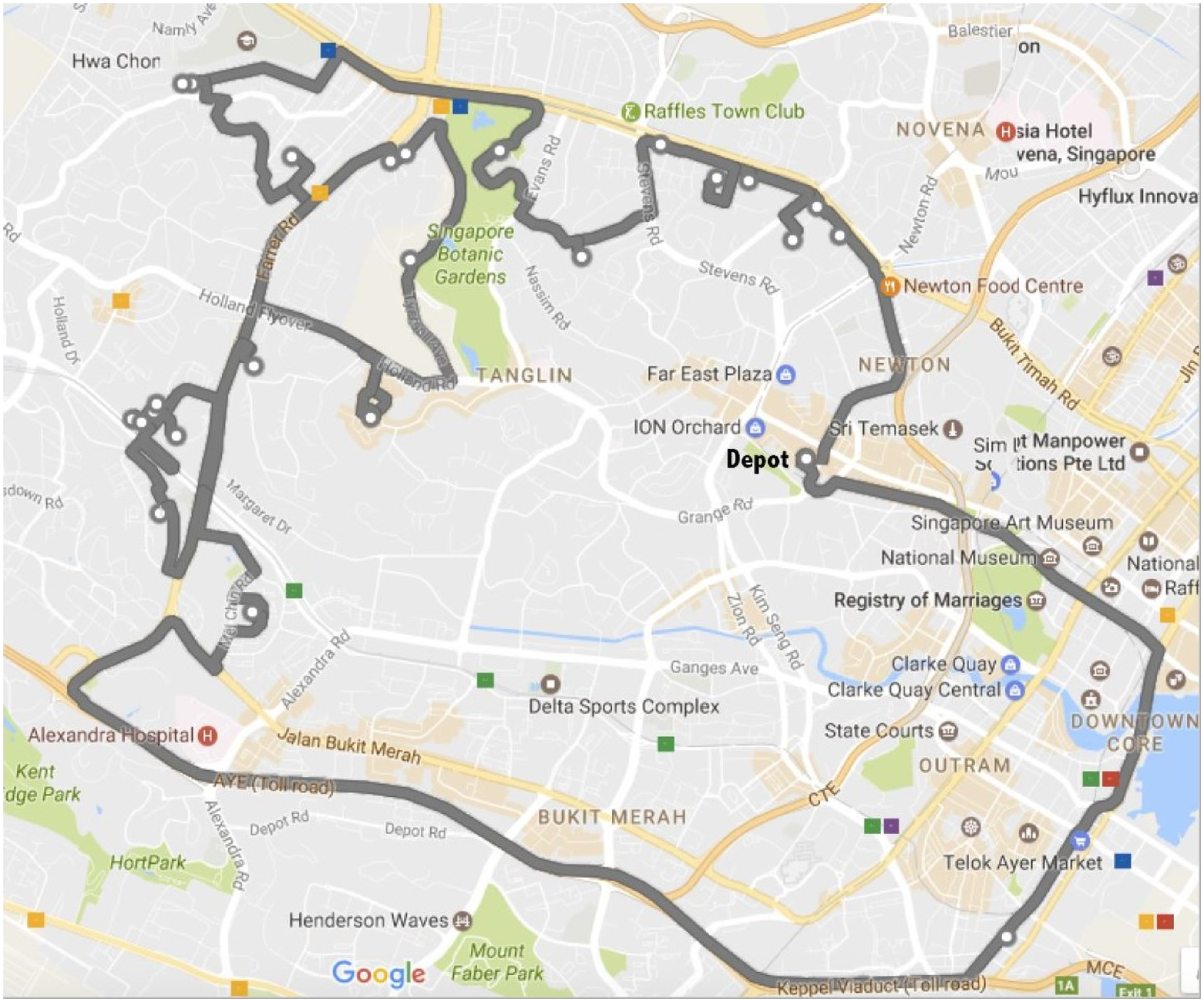} 
&\hspace{-1.3em}
 \includegraphics[width=0.39\textwidth]{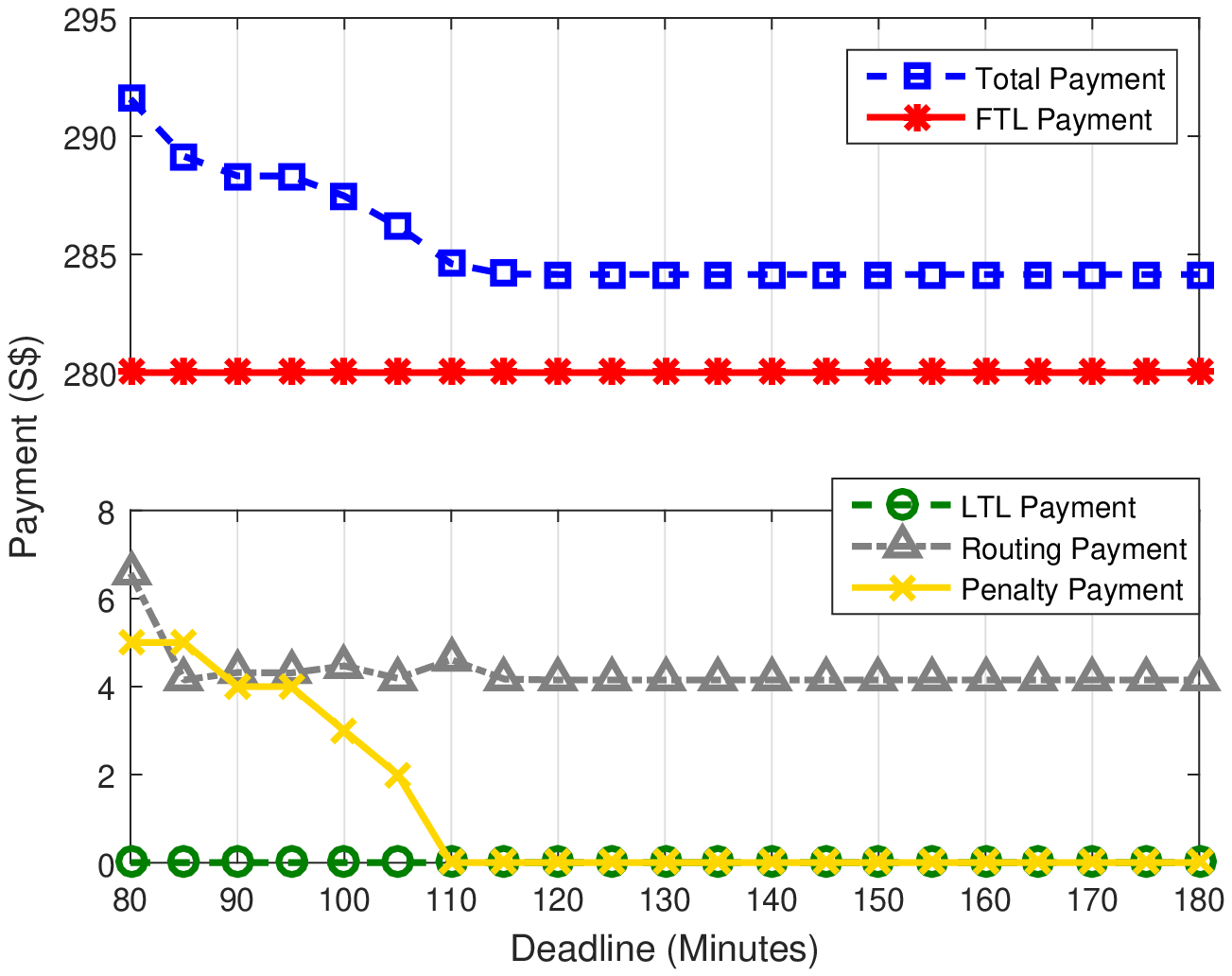} & \hspace{-2.7em}
 \includegraphics[width=0.39\textwidth]{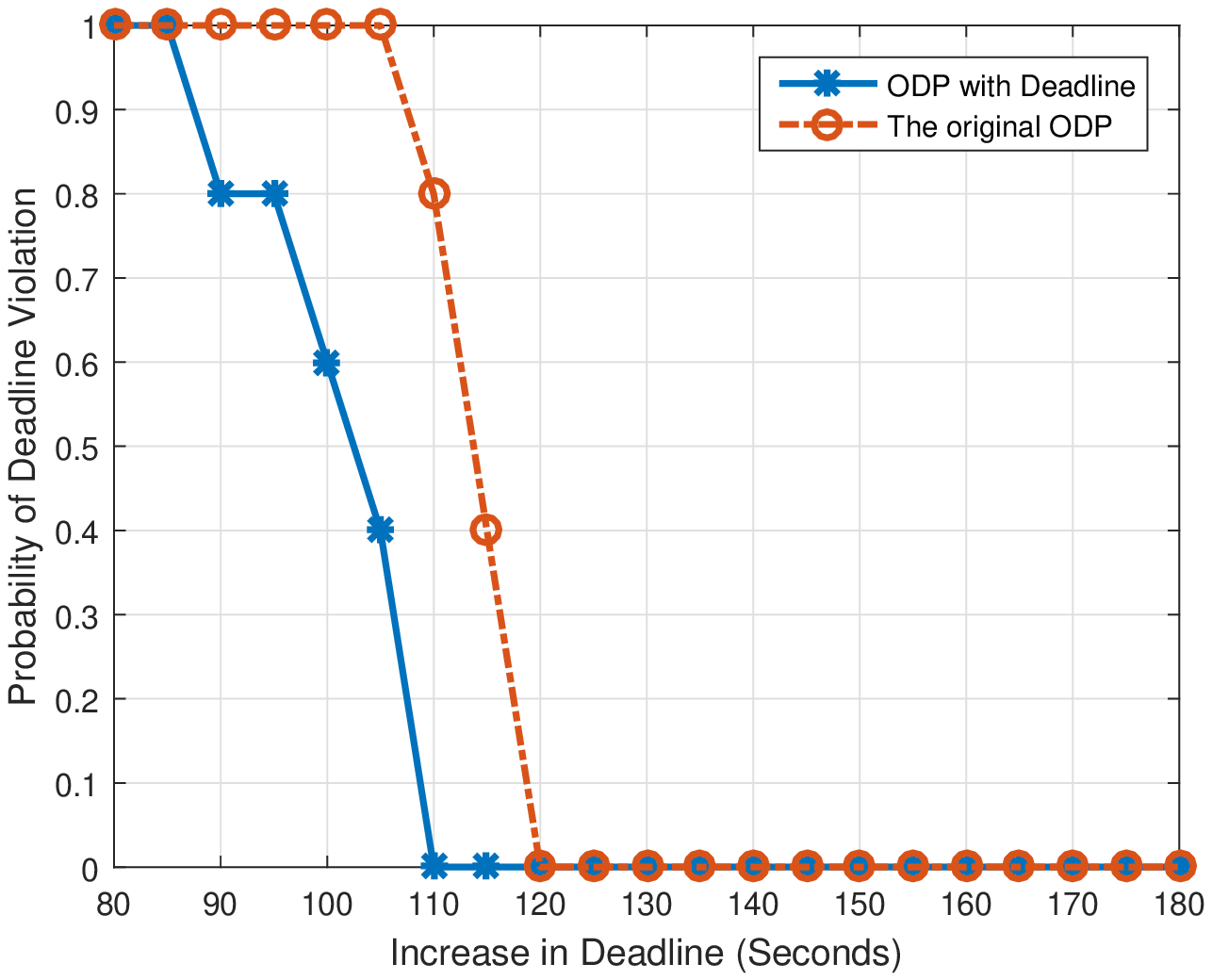} \\
\end{array}$

\minipage{0.3\textwidth}
 \caption{Vehicle routing for Singapore Road Map}
 \label{f_roadmap}
 \endminipage\hfill 
 \minipage{0.3\textwidth}
 \caption{Payment breakdown of different deadline when penalty cost~=~1}
 \label{f_deadline_cost}
 \endminipage\hfill
 \minipage{0.3\textwidth}
 \caption{Comparison between the original ODP and the ODPD on the deadline violation}
 \label{f_deadline_pro}
 \endminipage
\vspace{-1em}
$\begin{array}{ccc} 
\hspace{-3em}
\includegraphics[width=0.39\textwidth]{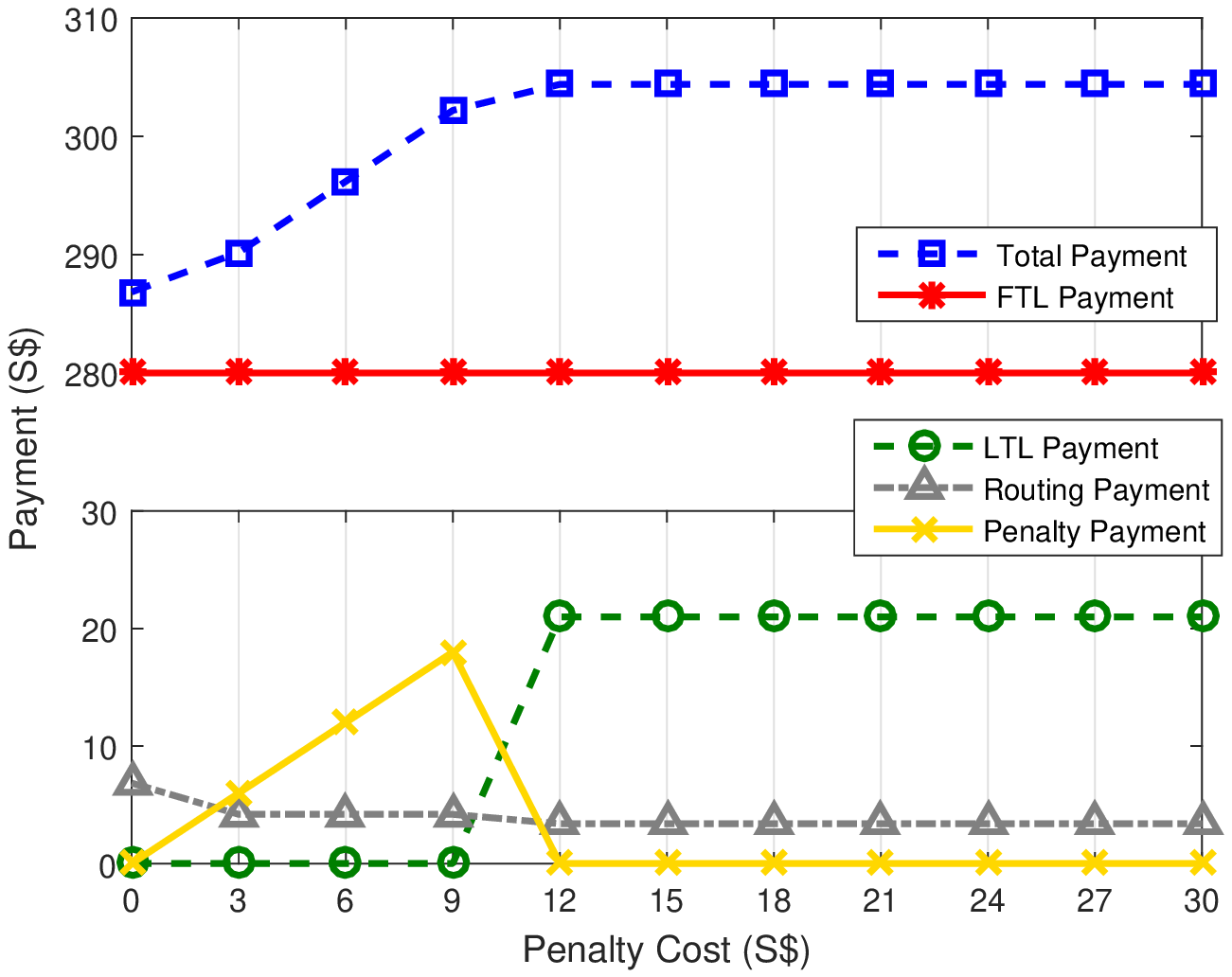} 
&\hspace{-2.55em}
 \includegraphics[width=0.39\textwidth]{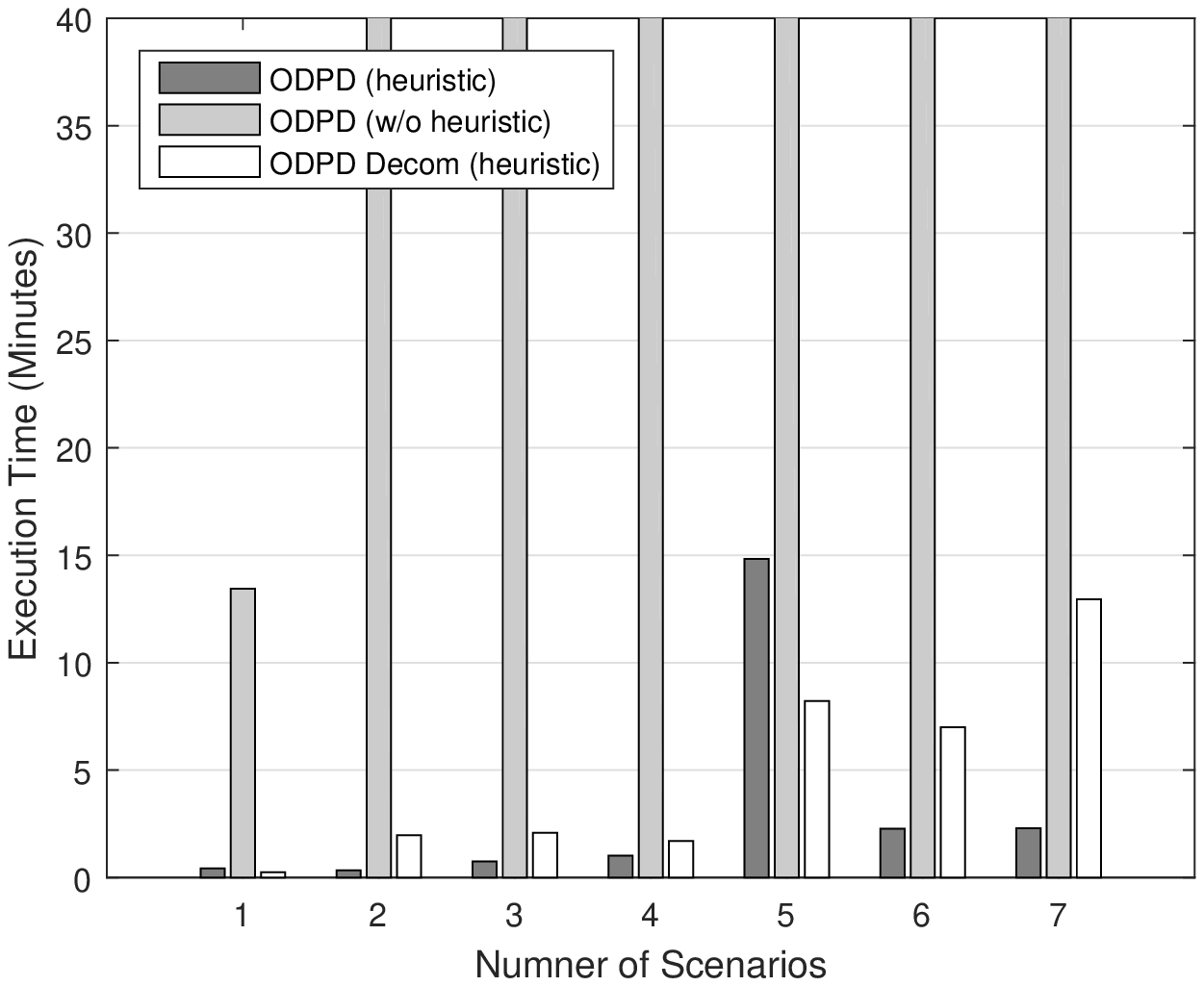} 
 & \hspace{-2.7em}
 \includegraphics[width=0.39\textwidth]{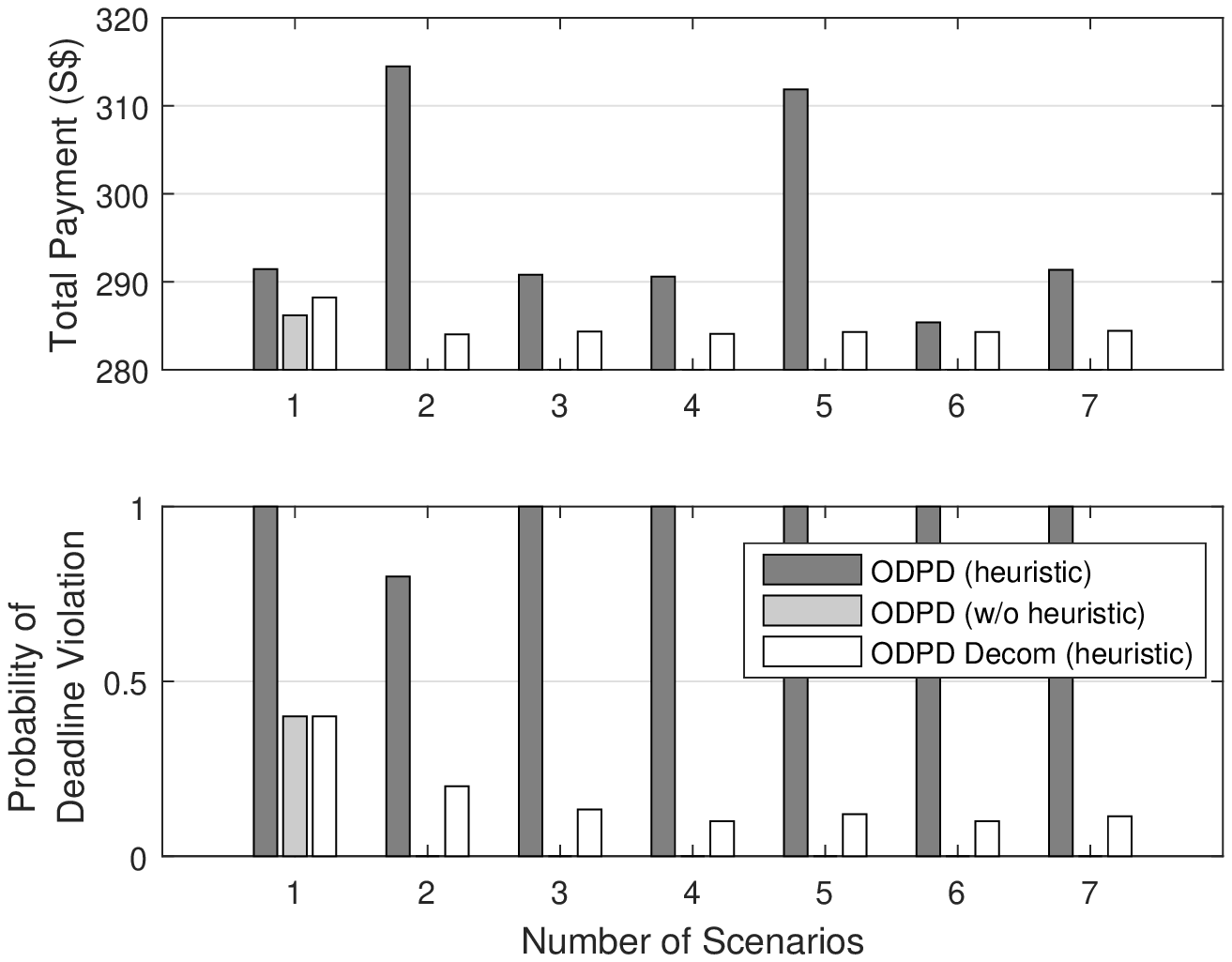} \\
\end{array}$
\minipage{0.3\textwidth}
 \caption{Different value of the penalty cost when the deadline = 105 minutes}
 \label{f_belta}
 \endminipage\hfill 
 \minipage{0.3\textwidth}
 \caption{Comparison of Time performance before and after decomposing}
 \label{f_decom_bar}
 \endminipage\hfill
 \minipage{0.3\textwidth}
 \caption{Comparison of Total Payment (top) and Deadline Violation Probability (bottom) before and after decomposing}
 \label{f_decom_line}
 \endminipage
 \vspace{-2em}
\end{figure*}

\section{Performance Evaluation}
\subsection{Parameter Setting}

We consider the system model with one FTL truck and one LTL carrier. The initial cost of a van, i.e., the FTL truck, is $\bar{C_{1}} = S\$280$, and the cost is estimated based on the truck renting price in Singapore~\cite{ref_sgvehicle}. The capacity limit of the van is $F_1 = 1060$ kilograms~\cite{ref_vehicle}. The routing cost between location $u$ to location $v$ is $\ddot{C}_{u,v} = K_{u,v} \times 0.105$, where $K_{u,v}$ is the distance in kilometers from location $u$ to location $v$~\cite{ref_gov}. In this study, we assume that all customer packages have equal weight in which $A = 30$ kilograms, and the cost of the LTL carrier is $\widehat{C} =S\$21$ per package. The cost of the LTL carrier is calculated based on the Speedpost service by SingPost Company~\cite{ref_singpost}. Note that the above parameter setting is similar to that in~\cite{ref_maggie_vtc}.

Moreover, the penalty cost is $\acute{C}$ = 1, and the deadline is $\mathbb{T}$ = 105 minutes. The error tolerance for terminating the decomposition loop  $\epsilon$ is 0.001. The customer demands are random in the cases of multiple demand scenarios, and all customers have demand in the case of one demand sceanrio.  Note that some parameters will be varied for different experiment. The ODPD system model is implemented in GAMS software~\cite{ref_gams}, the solver of which is CPLEX, to obtain an optimal solution. All the experiments are conducted on the PC computer with a 3.70GHz Intel Xeon CPU E5-1630 v3 processor and 32 GB installed memory (RAM).

In this paper, the input dataset is synthesized based on the real data from a logistics private limited company in Singapore and Google Map. We first use the geographical locations of the depot and 24 customers from the real logistics dataset. We implement a Java program to collect the distance and the traveling time of each customer's location from the Google Map Distance Matrix API~\cite{ref_googleAPI}. Additionally, to provide more variety of tests, we use the synthesized traveling time dataset which are generated from the normal distribution by using Google traveling time as the mean value, and the variance is set to be 100 seconds. 

\subsection{Routing Path and Payment}

The vehicle routing path is presented in Figure~\ref{f_roadmap} for the case that all 24 customers have demand. We observe that all the customers are served by the FTL truck. With the deadline of $\mathbb{T}$ = 105 minutes and the penalty cost of $\acute{C}$ = 1, the probability of deadline violation is 0.4. In other words, two out of five samples violate the deadline. Note that the total payment is S\$286.19, S\$280 of which is the initial cost of the FTL truck, S\$4.19 is the routing payment, and S\$2 is the penalty of violating deadline. Note that the penalty =  S\$2 because we ignore the denominator in the constraint in~(\ref{e_card_scount}).

\subsection{Impact of Different Deadline}

We vary the deadline for the following experiments while the penalty cost is fixed as one ($\acute{C} = 1$). For ease of presentation, we only consider the scenario that all customers have demand. When the deadline is set to be larger, there will be more choices for the system to select an optimal path without paying the penalty of deadline violation.

\subsubsection{Total Payment}

Figure~\ref{f_deadline_cost} shows the payment breakdown. When the deadline is between 80 and 85 minutes, no path can meet the deadline for all samples. For the deadline from 90 to 95 minutes, the system can find the optimal path with four samples which violate the deadline. When the deadline is set to 100 and 105 minutes,  three and two samples violate the deadline, respectively. The penalty payment reaches zero when the deadline is set to be 110 minutes, which means the system can find the feasible path without violating the deadline for all traveling time samples.

\subsubsection{Comparison of Deadline Violation Probability}

Figure~\ref{f_deadline_pro} shows the deadline violation probabilities of the original ODP and the ODPD. The probabilities are calculated from the optimal routing solutions. The ODPD always achieves smaller probability of deadline violation. This result clearly shows that the ODPD yields a better solution than that of the ODP as the ODPD explicitly incorporates the deadline violation into the formulation.

\subsection{Impact of Penalty Cost }

We consider the case that all customers have demand and the deadline is fixed (i.e., $\mathbb{T} = 105$ minutes). Then, we vary the value of the penalty cost. The payment breakdown is presented in Figure~\ref{f_belta}. When the penalty cost  $\acute{C} \leq 9$, all 24 customers are delivered by the FTL truck and the penalty cost increases as the penalty payment increases. However, when the penalty cost  $\acute{C} \geq 12$, which is too expensive compared with hiring the LTL carrier. Therefore, the ODPD system allocates 23 customers to the FTL truck and 1 customer to the LTL carrier. This solution guarantees that all the customers will be served before the deadline for all traveling time samples. Thus, the penalty payment is zero.

\subsection{Decomposition Evaluation}

We next evaluate the performance in terms of total payment and computation time of the ODPD with decomposition. We perform the following experiments by two different solver setting options, i.e., CPLEX with heuristic and CPLEX without a high speed feature or heuristic. The former may achieve a local optimal solution with fast computation. On the contrary, the latter is to obtain a global optimal solution. In general, the former executes faster than the latter. Note that the ODPD with decomposition is solved only using CPLEX with heuristic. The following results are obtained by varying the size of demand scenarios. We present only less than 7 scenarios in Figure~\ref{f_decom_bar} and  Figure~\ref{f_decom_line} because the  ODPD with heuristic does not yield the optimal solution  for more than 7 scenarios. In particular,  the customers are served by the outsourcing carrier, which can be considered as an unsolvable problem. Moreover, for multiple scenarios, the execution time of the ODPD without heuristic is too slow, i.e., taking more than 2 hours for 2 scenarios and 12 hours for 5 scenarios, which is unacceptable in practice.

The execution time of the ODPD is presented in Figure~\ref{f_decom_bar}. Note that we omit the results of  the ODPD without heuristic for multiple scenarios in Figure~\ref{f_decom_line} because the execution time is too long. For only one demand scenario case, the ODPD solved by CPLEX without heuristic achieves the best total payment (Figure~\ref{f_decom_line}~(top)). However, the total payment is less than S\$5 different compared with the other methods.

On the other hand, the ODPD and the decomposed ODPD, which are solved by CPLEX with heuristic, can be solved with acceptable execution time. 
The time performances of ODPD with and without decomposition are observed to be comparable as indicated in Figure~\ref{f_decom_bar}. 
However, the total payment and the deadline violation probability of the ODPD without decomposition, which is illustrated in Figure~\ref{f_decom_line}, are significantly higher than those of the decomposed ODPD. 

In summary, with the decomposition method, we can achieve an optimal solution for more than 7 scenarios with acceptable execution time. To illustrate, solving 25 scenarios using the decomposed ODPD is still faster than solving 2 scenarios using the original (undecomposed) ODPD without heuristic.  
The execution time of the decomposed ODPD increases according to number of scenarios. The decomposed ODPD always achieves the better total payment  than that of the ODPD when solving with CPLEX with heuristic.
Moreover, the deadline violation probability of the original (undecomposed) ODPD is always one or close to one which might not be preferable for the customers.


\section{Conclusion}

We have proposed the Optimal Stochastic Delivery Planning with Deadline (ODPD) to help supplier make the decision whether to serve customers by full-truckload (FTL) trucks or less-than-truckload (LTL) carriers before the actual customer demand is observed. The ODPD, which is able to accommodate the traveling time limit requirement,  enhances supplier benefit and customer satisfaction. The uncertainties in both customer demand and traveling time have been considered in the ODPD. While we assume that the distribution probability of the customer demand is known, the assumption of traveling time is not required as we have adopted the cardinality minimization approach in the ODPD routing. The L-shaped decomposition method has been applied to the ODPD. 
The experiments based on the real Singapore customer locations have been carried out. 
The decision between FTL truck and LTL carrier assignment has been optimized when the penalty cost of FTL trucks exceeding the deadline is accounted.  
The computational time of the decomposed ODPD is significantly lower than that of the original (undecomposed) ODPD. For the future work, multi-stage scenarios and FTL truck break down events will be incorporated.

\section{Acknowledgment}
This work is partially supported by Singapore Institute of
Manufacturing Technology-Nanyang Technological University
(SIMTech-NTU) Joint Laboratory and Collaborative research
Programme on Complex Systems.

\end{document}